\documentclass[twocolumn]{article}

\usepackage[width=17cm,height=22cm]{geometry}
\usepackage[english]{babel}
\usepackage[utf8]{inputenc}
\usepackage{fancyvrb}
\usepackage{authblk}
\usepackage{hyperref}
\usepackage{tikz}
\usepackage{amsmath}

\def\checkmark{\tikz\fill[scale=0.4](0,.35) -- (.25,0) -- (1,.7) -- (.25,.15) -- cycle;} 

\usepackage{times}
\usepackage{amssymb}
\usepackage[utf8]{inputenc}
\usepackage[T1]{fontenc}
\usepackage{graphicx}
\usepackage{hyperref}
\usepackage[rightcaption]{sidecap}

\usepackage{amsmath}
\usepackage[linesnumbered,ruled]{algorithm2e}

\title{Synapse at CAp 2017 NER challenge: Fasttext CRF}
\author[1,2]{Damien Sileo}
\author[1,*]{Camille Pradel}
\author[2,*]{Philippe Muller}
\author[2,*]{Tim Van de Cruys}

\affil[1]{Synapse Développement}
\affil[2]{IRIT, Université Paul Sabatier}
\affil[*]{Equal contributions}

\begin{document}
\maketitle

\begin{abstract}
We present our system for the CAp 2017 NER challenge \cite{Lopez2017} which is about named entity recognition on French tweets. Our system leverages unsupervised learning on a larger dataset of French tweets to learn features feeding a CRF model. It was ranked first without using any gazetteer or structured external data, with an F-measure of 58.89\%. To the best of our knowledge, it is the first system to use {\it fasttext} \cite{Bojanowski2006} embeddings (which include subword representations) and an embedding-based sentence representation for NER.
\end{abstract}

\medskip


\section{Introduction}

Named-Entity Recognition (NER) is the task of detecting word segments denoting particular instances such as  persons, locations or quantities. It can be used to ground knowledge available in texts.
While NER can achieve near-human performance \cite{Nrl1998}, it is is still a challenging task on noisy texts  such as tweets\cite{Ritter11} scarce labels, especially when few linguistic resources are available. Those difficulties are all present in the CAp NER challenge.

A promising approach is using unsupervised learning to get meaningful representations of words and sentences.
{\it Fasttext}  \cite{Bojanowski2006} seems a particularly useful unsupervised learning method for named entity recognition since it is based on the skipgram model which is able to capture substantive knowledge about words while incorporating morphology information, a crucial aspect for NER. We will describe three methods for using such embeddings along with a CRF sequence model, and we will also present a simple ensemble method for structured prediction (section \ref{sec:model}). Next, we will show the performance of our model and an interpretation of its results (section \ref{sec:results}). 

\section{Model}
\label{sec:model}

\begin{figure}[]
  \centering
\includegraphics[width=0.5\textwidth]{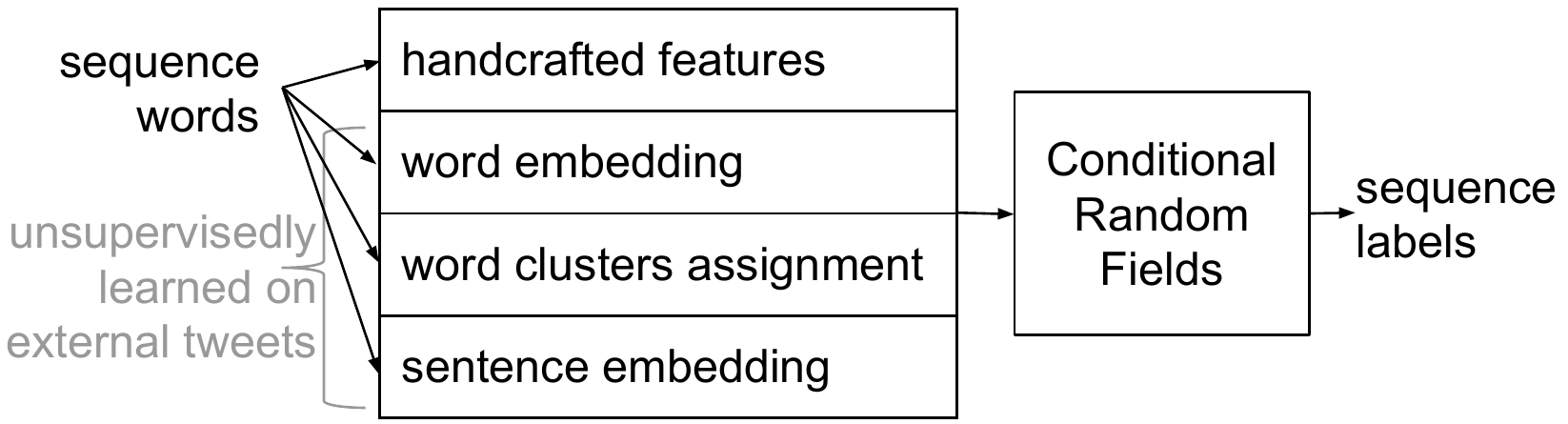}
  \caption{overview of our system}
\label{ov} 
\end{figure}

Figure \ref{ov} shows an overview of our model. This section will detail the components of the system. 
\subsection{CRF}
The core of our model is Conditional random fields (CRF) \cite{Sutton2011}, a structured prediction framework widely used in NER tasks. It can model the probabilities of a tag sequence $ y_1 ... y_ n  $ given a sequence of words $x_1 ...x_n$.

We use the linear chain CRF restriction where the sequences are modeled with the probability of transitions between consecutive labels.
\begin{equation}
P(y|x)= \frac{1}{Z(x)} 
\prod^n_{i=1} 
\text{exp}(\sum_j \theta_j f(y_{i-1}, y_{i}, x, i))
\end{equation}
$f$ yields a feature vector, $\theta$ is a weight vector, and $Z$ is a normalization factor in order to ensure a probability distribution. CRFs allow for non greedy optimization for learning sequence prediction and allows for much flexibility when defining the feature vector $f(y_{i-1}, y_{i}, x, i)$.
Furthermore, we can add a prior on the learned weights $\theta$ for regularization purposes. 
The likelihood of the training data can be optimized using gradient descent.
We chose $f$ to yield two sets of features that are concatenated: handcrafted features and {\it fasttext} embedding-based features.

\subsection{Handcrafted features}

Table \ref{hcf}  shows the handcrafted features we used. The context columns specifies whether or not a feature was also used with respect to the adjacent words.
\begin{table}[]
\centering
\caption{word-level handcrafted features}
\label{hcf}
\begin{tabular}{ll}
\textbf{feature}                       & \textbf{context} \\
word (lowercased)                      & \checkmark       \\
word length                            &                  \\
length 1 prefix                        & \checkmark       \\
length 2 prefix                        &                  \\
length 1 suffix                        & \checkmark       \\
length 2 suffix                        &                  \\
is\_upper                              & \checkmark       \\
is\_title                              & \checkmark       \\
position                               &                  \\
word uppercase proportion              &                  \\
word uppercase proportion*word length &                  \\
is\_emoji                              &                  \\
hyphenation                            &                  \\
POS tag                                &                  \\
is\_quote                              & \checkmark       \\
beginning of sentence                  &                  \\
end of sentence                        &                 
\end{tabular}
\end{table}

The emoji \footnote{https://pypi.python.org/pypi/emoji/} library was used for emoji detection, and we used the Treetagger \cite{Schmid1994} POS tagger.

\subsection{{\it Fasttext} features}

{\it Fasttext} skipgram is based on the {\it word2vec} skipgram model \cite{Mikolov2013}, where word representations are learned so that they optimize a task of predicting context words.
The main difference is that the representation  $h_w$ of a word $w$ is not only $u_w$, the representation of its symbol. It is augmented with the sum of the representations of its subword units $u_g, g \in \mathcal G_w$:
 \begin{equation}
h_w = u_w + \sum_{g  \in \mathcal G_w } u_g
\end{equation}
$\mathcal G_w$ encompasses some character n-grams that $w$ contains, provided they are frequent enough and of a desirable length.
Morphology of $w$ is thus taken in account in the representation of $h_w$ even though the order of n-grams is ignored.

$h_w$ can directly be used as a word level feature. However,  \cite{Guo2014} showed that CRFs work better with discrete features, so we also use a clustering-based representation.
Several approaches \cite{Ahmed2013,Sien2015,Das2017,Guo2014} use word embeddings for named entity recognition.

\subsubsection{Clustering {\it fasttext} features}
We cluster the {\it fasttext} representations of unique words in train and test tweets using a Gaussian Mixture Model (GMM), and feed the vector of probabilities assignments as word-level feature to the CRF. GMM clusters latent space to maximize the likelihood of the training data assuming that it is modeled by a mixture of gaussian.

\subsubsection{Sentence representation}
We also use the average of word representations in a tweet as a sentence level feature. It is a simple  way to provide a global context even though a linear model will not exploit this information thoroughly.

\subsection{Ensemble method}

We ensemble different models using a voting rule. We train $N$ systems, each time training an new {\it fasttext} model. This is the only variation between models, but different embeddings can influence the parameters learned with respect to handcrafted features. We then select the best prediction by picking the most frequent labeling sequence predicted for each tweet by the $N$ systems.
%
%
%
%
\section{Experimental settings}
\label{sec:exp}
Test/train data are from CAp NER 2017 data includes french labeled tweets with 13 kinds of segments and IOB format. Further details can be found in  \cite{Lopez2017}.
We used {\it Crfsuite}  \cite{Okazaki2007} through its {\it sklearn-crfsuite} python bindings \footnote{http://sklearn-crfsuite.readthedocs.io/en/latest/} which follows the {\it sklearn} API and allows for better development speed.
The original implementation of {\it fasttext} \cite{Bojanowski2006} was used through its python bindings. \footnote{https://github.com/salestock/fastText.py}

\subsection{Additional data}
To learn {\it fasttext} word representations, we used tweets from the {\it OSIRIM} \footnote{http://osirim.irit.fr/site/fr/articles/corpus} platform at IRIT, where  $1\%$ of the total feed of tweets is being collected since September 2015.
We picked a random subset of French tweets and dropped $99\%$ of tweets containing an url, since many of them come from bots. The remaining urls are kept because there are some urls in the challenge data. We replaced $1\%$ of mentions (\textit{@someone} tokens) by the symbol \textit{@*} hoping to help generalization.
This  preprocessed additional data totals 40M tweets.

\subsection{Parameter selection}

Parameters and feature subsets were not thoroughly optimized through cross validation, except regularization parameters.
We used Elasticnet regularization \cite{Zou2005} and the L-BFGS optimization algorithm, with a maximum of 100 iterations.\\
We ran grid search using sequence level accuracy score as a metric, on $c1$ and $c2$, the regularization weights for L1 and L2 priors. They were tested in respective ranges $[10^{-6}, 10^{-5}, 10^{-4}, 10^{-3}] $ and $[0.1,0.3, 0.8, 2]$.
$c1=10^{-4}$ and $c2=0.3$ were chosen.

{\it Fasttext} skipgram uses negative sampling with the parameters described in table \ref{ftparam}.
A different skipgram model was used for sentence representation, with a dimension of $40$. For the Gaussian Mixture Model, we used $100$ dimensions and diagonal covariance.

\begin{table}[]
\centering
\caption{{\it Fasttext} parameters}
\label{ftparam}
\begin{tabular}{ll}
\textbf{parameter}              & \textbf{value} \\
learning rate                   & 0.02           \\
dimension                       & 200            \\
context window size             & 5              \\
number of epochs                & 4              \\
min\_count                      & 5              \\
negative/positive samples ratio & 5             \\
minimum n-gram size             & 3              \\
maximum n-gram size             & 6              \\
sampling threshold              & $10^{-4}$          
\end{tabular}
\end{table}

\section{Results}
\label{sec:results}
\subsection{Clustering}
Many clusters correspond directly to named entities. Table \ref{clusters} shows a random sample of 10 handpicked clusters and the themes we identified.
\begin{table}[]
\centering
\caption{Handpicked clusters and random samples}
\label{clusters}
\begin{tabular}{ll}
\textbf{Cluster theme} & \textbf{Cluster sample}                                                                                                             \\
hyperlinks                    & https://t.co/d73eViSrbW \\
hours                  & 12h00 19h19 12h 7h44                                                                                 \\
dates                  & 1947 1940  27/09 mars Lundi                                                                       \\
joyful reactions        &  ptdrr mdrrrrrrr pff booooooooordel                                                         \\
TPMP (french show)     &  \#TPMP \#hanouna Castaldi                              \\
transportation lines   & @LIGNEJ\_SNCF @TER\_Metz\\
emojis                 &       Pfffff   :)                                                                                                 \\
video games            & @PokemonFR manette RT       \\
persons                & @olivierminne @Vibrationradio  \\
football players       & Ribery Leonardo Chelsea Ramos                                                 
\end{tabular}
\end{table}

\subsection{Performance}

We will report the results of the system on the evaluation data when fitted on the full training data.
The system yields a sequence level accuracy of $61.2 \%$ using an ensemble of $N=20$ models. Note that a single model ($N=1$) has a sequence level accuracy $60.7 \%$ which is only slightly less.

The challenge scoring metric was a micro F-measure based on chunks of consecutive labels. Our ensemble system scores  58.89\% with respect to this metric. Table \ref{scores} summarize the results of the competition and show that our system won with a rather large margin.
%
%
{\it Fasttext} features bring a notable difference since the sequence level accuracy drops to $57.8\%$ when we remove all of  them.
Table \ref{detailedscores} gives an overview of scores per label, and could show us ways to improve the system. The 13 labels were separated according to their IOB encoding status.\\

\begin{table}[]
\centering
\caption{Fine grained score analysis}
\label{detailedscores}
\begin{tabular}{lllll}
label           & precision & recall & f1-score & support \\
B-person        & 0.767     & 0.618  & 0.684    & 842     \\
I-person        & 0.795     & 0.833  & 0.814    & 294     \\
B-geoloc        & 0.757     & 0.697  & 0.726    & 699     \\
B-transportLine & 0.978     & 0.926  & 0.951    & 517     \\
B-musicartist   & 0.667     & 0.178  & 0.281    & 90      \\
B-other         & 0.286     & 0.134  & 0.183    & 149     \\
B-org           & 0.712     & 0.277  & 0.399    & 545     \\
B-product       & 0.519     & 0.135  & 0.214    & 312     \\
I-product       & 0.320     & 0.113  & 0.167    & 364     \\
B-media         & 0.724     & 0.462  & 0.564    & 210     \\
B-facility      & 0.639     & 0.363  & 0.463    & 146     \\
I-facility      & 0.620     & 0.486  & 0.545    & 175     \\
B-sportsteam    & 0.514     & 0.277  & 0.360    & 65      \\
I-sportsteam    & 1.000     & 0.200  & 0.333    & 10      \\
B-event         & 0.436     & 0.185  & 0.260    & 92      \\
I-event         & 0.356     & 0.292  & 0.321    & 89      \\
B-tvshow        & 0.429     & 0.058  & 0.102    & 52      \\
I-tvshow        & 0.286     & 0.065  & 0.105    & 31      \\
I-media         & 0.200     & 0.019  & 0.035    & 52      \\
B-movie         & 0.333     & 0.045  & 0.080    & 44      \\
I-other         & 0.000     & 0.000  & 0.000    & 73      \\
I-transportLine & 0.873     & 0.729  & 0.795    & 85      \\
I-geoloc        & 0.650     & 0.409  & 0.502    & 159     \\
I-musicartist   & 0.636     & 0.163  & 0.259    & 43      \\
I-movie         & 0.250     & 0.049  & 0.082    & 41     
\end{tabular}
\end{table}

\subsection{Interpreting model predictions}

CRF is based on a linear model and the learned weights are insightful:
the highest weights indicate the most relevant features for the prediction of a given label, while the lowest weights indicate the most relevant features for preventing the prediction of a given label.
Tables \ref{hw} and \ref{lw} show those weights for a single model trained on all features.
{\it ft\_wo\_i}, {\it ft\_wo\_c\_i} and {\it ft\_sen\_i} refer respectively to the $i$th component of a {\it fasttext} raw word representation, cluster based representation, and sentence level representation.
The model actually uses those three kinds of features to predict labels.
Clustering embeddings can improve the interpretability of the system by linking a feature to a set of similar words. Sentence level embeddings seem to prevent the model from predicting irrelevant labels, suggesting they might help for disambiguation.

\begin{table}[]
\centering
\caption{Highest $\theta$ weights}
\label{hw}
\begin{tabular}{lll}
\textbf{weight} & \textbf{label}    & \textbf{feature}            \\
3.26            & O               & end of sentence                         \\
2.47            & O               & beginning of sentence                        \\
2.01            & O               & previous word:rt            \\
1.92            & B-transportLine & ft\_wo\_91                    \\
1.85            & B-other         & previous word:les           \\
1.80            & B-geoloc        & previous word:\#qml         \\
1.76            & B-geoloc        & previous word:pour          \\
1.71            & B-geoloc        & ft\_sen\_22                 \\
1.71            & O               & ft\_wo\_c68                 \\
1.68            & B-org           & current word:\#ratp         \\   
\end{tabular}
\end{table}

\begin{table}[]
\centering
\caption{Lowest  $\theta$ weights}
\label{lw}
\begin{tabular}{lll}
\textbf{weight} & \textbf{label}    & \textbf{feature}            \\
-1.65           & B-product       & ft\_sen\_33            \\
-1.60           & B-org           & ft\_sen\_9                  \\
-1.48           & O               & previous word:sur           \\
-1.41           & B-facility      & ft\_sen\_33                 \\
-1.40           & O               & suffix:lie                  \\
-1.38           & O               & suffix:ra                   \\
-1.29           & B-other         & previous POS: verb (future) \\
-1.29           & B-geoloc        & ft\_wo\_151                   \\
-1.27           & B-person        & previous word prefix:l      \\
-1.26           & B-org           & ft\_wo\_130                   \\
\end{tabular}
\end{table}

\subsection{Computational cost}
Fitting the CRF model with 3000 examples (labeled tweets) takes up 4 minutes on a Xeon  E5-2680 v3 CPU  using a single thread, and inference on 3688 example only needs 30 seconds.
Fitting the {\it fasttext} model of dimension 200 on 40M tweets takes up 10 hours on a single thread, but only 30 minutes when using 32 threads.

\section{Conclusion and further improvements}

We presented a NER system using {\it Fasttext} which was ranked first at the CAP 2017 NER challenge.
Due to a lack of time, we did not optimize directly on the challenge evaluation metrics, using sequence level accuracy as a proxy, and we did not cross-validate all important parameters. 
%
%
Besides, there are  other promising ways to increase the score of the system that we did not implement:
\begin{enumerate}

\item  thresholding for F1 maximization: Our system precision ($73.65 \%$) is significantly higher than its recall ($49.06 \%$). A more balanced score could be obtained by having a negative bias towards predicting no label. This might  improve the F1 score. Threshold optimization works well for non-structured prediction \cite{ChaseLipton2014}, but it is not clear that it would bring about improvement in practical applications.
\item  larger scale unsupervised learning: More tweets could be used, and/or domain adaptation could be applied in order to bias embeddings towards learning representations of words occurring in the challenge data.
\item  RNN embeddings: Unsupervised learning with recurrent neural networks can be used to learn "contextualized" embedding of words. Unsupervised training tasks include language modeling or auto-encoding. RNNs have been used in NER without unsupervised training. \cite{DBLP:journals/corr/AthavaleBPPS16} \cite{Limsopatham2016}
\item   DBPedia spotlight \cite{Daiber2013} could provide an off-the-shelf gazetteer, yielding potentially  powerful features for NER.
\end{enumerate}

\bibliography{biblio12}

\end{document}